%% file: TeleDex.tex
\documentclass[11pt, letterpaper]{article}
\input{preamble.tex} 

\title{\textbf{Technical Report: \\ 
Open TeleDex: A Hardware-Agnostic Teleoperation System for Imitation Learning based
Dexterous Manipulation}}

\author[a,1]{Xu Chi}
\author[a,1]{Chao Zhang}
\author[c,1]{Yang Su}
\author[b]{Lingfeng Dou}
\author[a]{Fujia Yang}
\author[a]{Jiakuo Zhao}
\author[c]{Haoyu Zhou}
\author[c]{Xiaoyou Jia}
\author[c]{Yong Zhou}   
\author[a,*]{Shan An} 
\affil[a]{Tianjin University}
\affil[b]{Tsinghua University}
\affil[c]{LinkerBot Inc.}  
\affil[1]{These authors contributed equally to this work.}
\affil[*]{Corresponding author: \texttt{anshan@tju.edu.cn}}
\date{\today}

\begin{document}

\maketitle

\begin{abstract}
Accurate and high-fidelity demonstration data acquisition is a critical bottleneck for deploying robot Imitation Learning (IL) systems, particularly when dealing with heterogeneous robotic platforms. Existing teleoperation systems often fail to guarantee high-precision data collection across diverse types of teleoperation devices. To address this, we developed Open TeleDex, a unified teleoperation framework engineered for demonstration data collection. Open TeleDex specifically tackles the TripleAny challenge, seamlessly supporting any robotic arm, any dexterous hand, and any external input device. Furthermore, we propose a novel hand pose retargeting algorithm that significantly boosts the interoperability of Open TeleDex, enabling robust and accurate compatibility with an even wider spectrum of heterogeneous master and slave equipment. Open TeleDex establishes a foundational, high-quality, and publicly available platform for accelerating both academic research and industry development in complex robotic manipulation and IL. 
\end{abstract}

\section{Introduction}
\label{sec:introduction}

Executing a wide spectrum of complex, dexterous manipulation tasks in diverse, unstructured environments remains a central challenge in modern robotics. Although Imitation Learning (IL) has shown great potential in tackling this problem, its progress is fundamentally constrained by the difficulty of collecting large-scale, multi-modal, and high-quality demonstration data~\cite{brohan2022rt1}.

The quality and scalability of this data hinge entirely on the underlying teleoperation system. These systems are designed around a series of complex trade-offs, often balancing three core dimensions: Fidelity (the precision of control and data quality), Accessibility (cost and deployment complexity), and Generalization (hardware compatibility). Current solutions consistently compromise on at least one dimension. For instance, high-fidelity systems, such as dedicated master-slave arms, offer unmatched precision and minimal latency, making them indispensable in zero-tolerance applications. However, this tight electromechanical coupling inherently limits their Generalization, as they exist as costly, "closed ecosystems" tailored to a single robot model, making cross-platform data aggregation impractical. Conversely, systems prioritizing Accessibility by leveraging consumer-grade VR/AR devices have significantly reduced the technical barriers to data collection. Despite this, they frequently compromise on Fidelity due to challenges like visual tracking uncertainties, communication latency, and the absence of physical force feedback. Furthermore, even these more accessible platforms often require substantial software adaptation and parameter tuning when faced with a new robotic arm or hand, revealing a critical generalization gap in practice. This lack of a unified framework supporting heterogeneous robotic platforms across the entire data collection workflow is the core bottleneck hindering the evolution of IL research prototypes into scalable, general-purpose data collection platforms.

To address this, we introduce Open TeleDex, a device-agnostic and highly versatile teleoperation framework. Our core design tenet is Generalization, achieved through a natively ROS2-based architecture that intrinsically unifies diverse control paradigms, allowing for the rapid and seamless integration of new hardware combinations. Our work makes two core contributions:

\begin{enumerate}
    \item \textbf{A Unified, Hardware-Agnostic Teleoperation Framework:} We present a decoupled, three-tier architecture that supports synchronous data collection from a wide variety of robotic arms, multi-fingered dexterous hands, and external input devices. Open TeleDex serves as an extensible backbone, allowing researchers to efficiently integrate and switch between diverse heterogeneous hardware combinations.
    
    \item \textbf{A Hand Pose Retargeting Algorithm:} We propose a novel retargeting algorithm that adjusts for kinematic discrepancies between human and robot hands. Our algorithm is designed to enhance the framework’s compatibility.
\end{enumerate}

This report is organized as follows: Section 2 the reviews related work. Section 3 details Open TeleDex framework. Section 4 shows the evaluation of our system. Finally, Section 5 concludes the report and discusses future work.


\section{Related Work}

\subsection{Teleoperation Systems}

Teleoperation systems are a crucial tool for collecting expert demonstrations for imitation learning, and their design inherently involves a series of complex trade-offs. The efficacy of any such system can be evaluated along three core dimensions. The first is Fidelity, which measures the precision with which a system conveys operator intent and relays environmental feedback; this remains a central focus in modern telerobotics research~\cite{hulin2011fidelity}. The second is Accessibility, encompassing its cost, deployment complexity, and ease of use, which are critical factors for practical application~\cite{nielsen1994usability}. The third, and increasingly critical dimension, is Generalization—the system's capacity to adapt to diverse robot hardware, tasks, and environments, a key challenge in modern human-robot interaction~\cite{goodrich2007human}. Existing solutions in the field typically exhibit distinct strengths and compromises across these dimensions.

Systems prioritizing ultimate fidelity, such as dedicated master-slave arms, have set the gold standard for latency and haptic "transparency", making them indispensable in zero-tolerance applications like robotic surgery~\cite{lee2018fitts}.  Pioneering systems like ALOHA~\cite{zhao2023aloha} have democratized data collection by drastically reducing the financial and technical barriers to entry for dexterity research. Through tight electromechanical coupling and direct joint-space mapping, they minimize information loss during transmission. However, this design philosophy inherently constrains their generalization. They often exist as "closed ecosystems" tailored to a single robot model, where any hardware modification can demand costly redesigns, rendering cross-platform deployment and data aggregation impractical.

In recent years, a wave of research has focused on improving accessibility by leveraging consumer-grade VR/AR devices and standard cameras. These systems typically employ more flexible Cartesian-space control. Despite their success in accessibility, they often compromise on fidelity, facing challenges from visual tracking uncertainties, communication latency, and the absence of physical force feedback~\cite{sivakumar2022robotic, qin2022from}. More importantly, while more general in theory, many open-source implementations still require substantial software adaptation and parameter tuning for specific robots, revealing a generalization gap in practice.

It is thus evident that generalization has emerged as the core bottleneck connecting the worlds of high fidelity and high accessibility, and hindering the evolution of teleoperation technology from one-off research prototypes to scalable data collection platforms. Addressing this challenge necessitates the development of a universal software framework. Works like AnyTeleop~\cite{qin2023anyteleop} have made important strides in this direction with learning-free, GPU-accelerated components. However, there remains a need for a framework that is natively rooted in the ROS2 ecosystem and is architected to intrinsically unify diverse control paradigms, from direct mapping to planner-centric control.

The TeleDex framework presented in this paper aims to build such a unified teleoperation framework with generalization as its core design principle. It does not seek to invent a new interaction modality, but rather to integrate and empower the diverse existing hardware ecosystem by providing a modular, extensible software backbone, thereby offering a truly scalable data collection solution for robot imitation learning.

\subsection{Hand Pose Retargeting}

The real-time and accurate transfer of a human operator's natural movements to a morphologically different robot is a core challenge in general-purpose teleoperation. This challenge can be decomposed into two main sub-problems: arm motion generation and dexterous hand pose retargeting.

For arm motion generation, mainstream approaches rely on Inverse Kinematics (IK) solvers or more advanced real-time motion planners~\cite{corke2017robotics, qin2023anyteleop}. These methods all require a precise kinematic model of the robot (e.g., a URDF file) to compute joint trajectories that match the operator's end-effector pose intent.

Dexterous hand posture retargeting is a more complex challenge. Existing methods, whether optimization-based~\cite{qin2023anyteleop, arunachalam2022holodex} or learning-based~\cite{li2019vision, zhang2021human}, mostly follow an analytical paradigm: they pre-define keypoint or joint correspondences between the human and robot hands and then minimize the error between them. While intuitive, the fundamental limitation of this "point-to-point mapping" approach lies in its rigid correspondence. When the robot hand's morphology differs significantly from a human's, this mapping can easily produce unnatural or physically infeasible postures, and the process of configuring these correspondences for new hardware is tedious.

To overcome the limitations of this "rigid mapping" paradigm, we propose a novel "generative" hand retargeting algorithm. Instead of copying each joint motion, our method treats the robot hand as an integrated skeleton constrained by its own kinematic model. Our algorithm drives this skeleton to "grow" naturally to functionally match the human's overall grasping intent, rather than mimicking it point-by-point morphologically. This shift from "copying motion" to "reproducing function" enables our framework to generate more natural and functional postures for morphologically diverse dexterous hands.

\subsection{Demonstration Data for Imitation Learning}

The success of Imitation Learning (IL) is critically dependent on the quality of demonstration data~\cite{osa2018algorithmic}. Research has shown that high-quality demonstrations, characterized by smooth motions and precisely synchronized state-action pairs, can significantly improve the final policy's performance and robustness~\cite{rajeswaran2018learning, peng2018deepmimic}. Therefore, building a collection framework that systematically ensures data quality is essential.

In practice, ensuring data quality faces two main challenges. The first is time synchronization. In a system with heterogeneous sensors like cameras and encoders, ensuring a common and precise time base for all data streams is a core difficulty. Simple recording schemes can easily introduce temporal misalignments that are detrimental to learning dynamic tasks~\cite{lynch2020latent}. The second challenge is the richness of data modality. While traditional master-slave systems provide high-quality kinematic data, modern IL increasingly benefits from multi-modal information, including multi-view images and force/tactile feedback~\cite{wang2024dexcap}. Emerging paradigms like VR teleoperation (e.g., Holo-Dex~\cite{arunachalam2022holodex}) are gaining attention for their natural ability to fuse multi-modal data.

This necessitates a modern data collection framework that not only features high-performance real-time processing and precise time synchronization but also natively supports the fusion of various collection methods in its architecture to meet the growing demand for high-quality, multi-modal, and diverse data.

\section{Framework of Open TeleDex}

\begin{figure}[!t]
    \centering
    \includegraphics[width=1.0\textwidth]{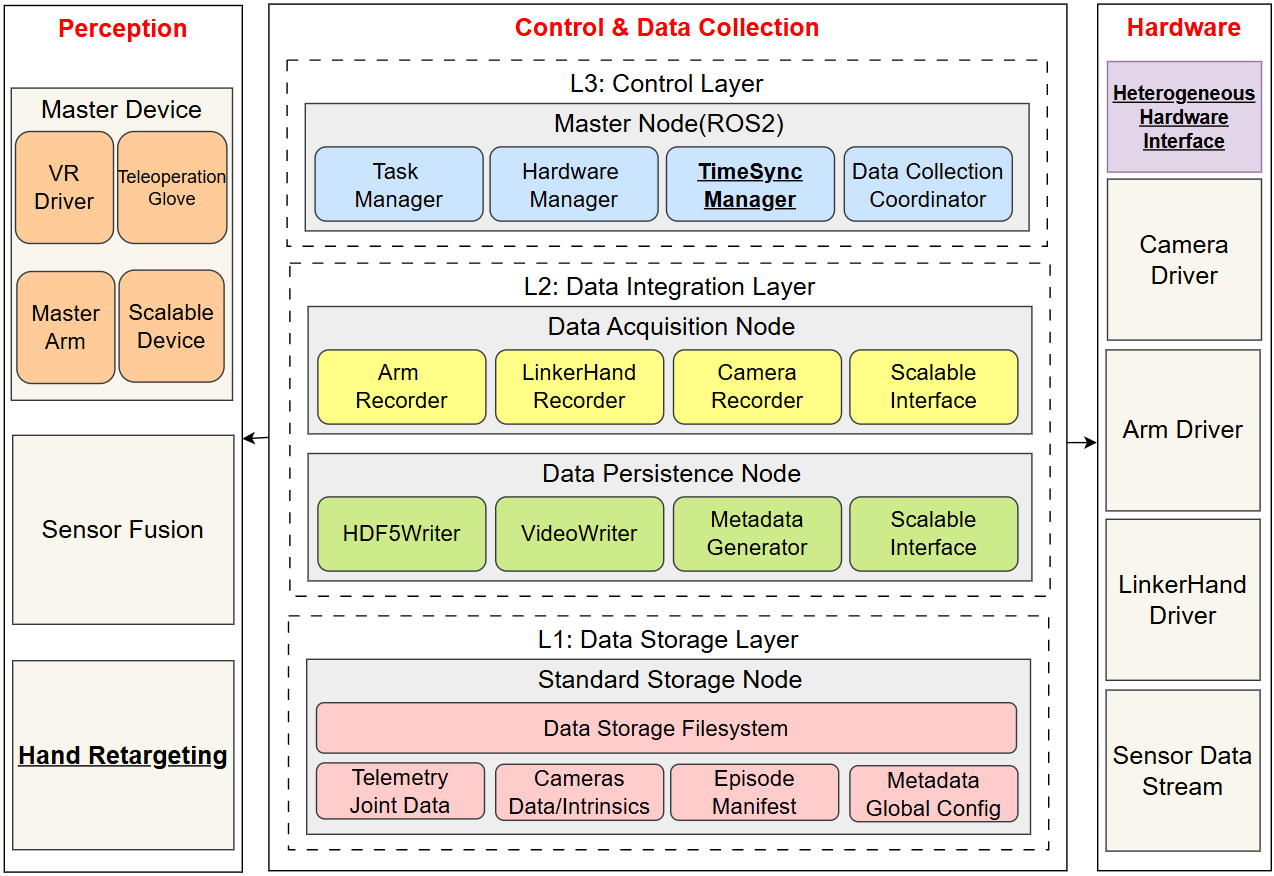} 
    \caption{The Framework of Open TeleDex. The diagram illustrates the layered architecture of the system, which is designed to support heterogeneous robot platforms for different requirement of data collection. The system is structurally divided into three modules: Perception, Control and Data Collection, and Hardware.}
    \label{fig:techsys}
\end{figure}

\subsection{System Architecture}

As shown in Fig.~\ref{fig:techsys}, the architecture of Open TeleDex follows the core design principles of layering and decoupling, with the ultimate goal of realizing our \textbf{TripleAny} vision: supporting the seamless integration and cooperation of \textbf{AnyExternalDevice}, \textbf{AnyArm}, and \textbf{AnyHand}. It primarily utilizes the ROS2 framework~\cite{macenski2022ros2} for distributed communication. Conceptually, the architecture is divided into three core modules:

\begin{enumerate}
    \item \textbf{Perception}, responsible for standardizing heterogeneous input data from Any External Device.
    
    \item \textbf{Control \& Data Collection}, which executes hardware-independent task management, control strategy scheduling, and data processing. This core layer is "agnostic" to the specific arm and hand, which is key to achieving AnyArm and AnyHand.
    
    \item \textbf{Hardware}, which translates standardized control commands for specific AnyArm and AnyHand actuators.
\end{enumerate}

\subsection{Perception}
The Perception component is responsible for capturing and interpreting the operator's intent and movement, forming the input signals for the entire teleoperation system.

Master Device: this module encapsulates various human interface devices such as Virtual Reality (VR) headsets (for spatial tracking), Teleoperation Gloves (for finger articulation capture), and Master Arms.

Sensor Fusion: Raw data from disparate input sensors is fused and filtered to generate a coherent, low-latency stream of the operator's hand and arm pose.

\paragraph{Hand Pose Retargeting:} this is a high-level decision module responsible for transforming the continuous human pose stream into a discrete, feasible goal pose for the robotic Slave devices. To achieve precise and adaptable teleoperation for robotic hands, we construct a modular optimization framework. The core of this framework is a total cost function, which is a weighted sum of three distinct sub-functions: a pose alignment cost, an inter-finger coupling cost, and a temporal smoothness cost. By minimizing this total cost function, we can solve for the optimal robot joint angle vector, $\theta^*(t)$, in real-time.

\textbf{Optimization Objective:} The optimal robot joint angle vector, $\theta^*(t)$, is found by solving the following minimization problem:
\begin{equation}
    \theta^*(t) = \arg\min_{\theta(t)} \left( \alpha_{\text{align}}\mathcal{L}_{\text{align}}(\theta(t)) + \alpha_{\text{couple}}\mathcal{L}_{\text{couple}}(\theta(t)) + \alpha_{\text{smooth}}\mathcal{L}_{\text{smooth}}(\theta(t)) \right)
    \label{eq:total_loss}
\end{equation}
Where:
\begin{itemize}
    \item $\theta(t) \in \mathbb{R}^n$ is the robot hand's joint angle vector at time $t$.
    \item $\mathcal{L}_{\text{align}}$, $\mathcal{L}_{\text{couple}}$, and $\mathcal{L}_{\text{smooth}}$ represent the three core cost functions.
    \item $\alpha_{\text{align}}$, $\alpha_{\text{couple}}$, and $\alpha_{\text{smooth}}$ are hyperparameters (weights) that balance the importance of each cost term.
\end{itemize}

\textbf{Cost Function:} \textit{Pose Alignment Cost ($\mathcal{L}_{\text{align}}$)}. 
This cost function aims to minimize the geometric error between the operator's hand keypoints and the corresponding keypoints on the robot hand. It is central to achieving the fundamental pose mimicry.
\begin{equation}
    \mathcal{L}_{\text{align}}(\theta(t)) = \sum_{(i,j) \in \mathcal{K}} \| p'_{i,j}(t) - \mathcal{F}_{k,j}(\theta(t)) \|^2
    \label{eq:align_loss}
\end{equation}
Here, $p'_{i,j}(t) \in \mathbb{R}^3$ represents the position of the operator's $j$-th keypoint on the $i$-th finger after transformation and calibration, while $\mathcal{F}_{k,j}(\theta(t))$ is the robot's forward kinematics function, which calculates the position of its corresponding keypoint.

\textit{Operator Keypoint Transformation.}
To compute the transformed keypoints $p'_{i,j}(t)$, we must first scale and translate the raw detected keypoints, $p_{i,j}(t)$, to match the robot's geometry.
\begin{enumerate}
    \item Dimensional Scaling Factor ($s_{i,j}$): This calculates the relative length ratio for each phalangeal segment.
    \begin{equation}
        s_{i,j} = \frac{\| \mathcal{F}_{k,j+1}(\theta_0) - \mathcal{F}_{k,j}(\theta_0) \|}{\| \bar{p}_{i,j+1} - \bar{p}_{i,j} \|}
    \end{equation}
    Here, $\bar{p}_{i,j}$ and $\theta_0$ represent the human keypoints and robot joint angles from a static calibration pose, respectively.
    
    \item Finger Root Translation Vector ($\delta_i$): This calculates the offset for the base position of each finger.
    \begin{equation}
        \delta_i = \mathcal{F}_{k,i,\text{mcp}}(\theta_0) - \bar{p}_{i,\text{mcp}}
    \end{equation}
    
    \item Transformed Keypoint ($p'_{i,j}$): Finally, the transformed points are obtained via the following iterative formula:
    \begin{equation}
    p'_{i,j}(t) =
    \begin{cases}
        p_{i,0}(t), & j=0 \\
        p'_{i,j-1}(t) + s_{i,j-1}(p_{i,j}(t) - p_{i,j-1}(t)) + \delta_i, & j=1 \\
        p'_{i,j-1}(t) + s_{i,j-1}(p_{i,j}(t) - p_{i,j-1}(t)), & j \geq 2
    \end{cases}
    \end{equation}
\end{enumerate}

\textit{Inter-Finger Coupling Cost ($\mathcal{L}_{\text{couple}}$).}
This cost function is used to maintain human-like coordination between the thumb and other fingers during fine motor actions such as grasping and pinching.
\begin{equation}
    \mathcal{L}_{\text{couple}}(\theta(t)) = \sum_{i \in \mathcal{I}} \beta_i(t) \| \vec{R}_{H,i}(t) - \vec{R}_{R,i}(\theta(t)) \|^2
    \label{eq:couple_loss}
\end{equation}
Here, $\vec{R}_{H,i}(t) = p_{i,\text{fingertip}}(t) - p_{\text{thumb,fingertip}}(t)$ is the relative vector from the operator's thumb tip to other fingertips. $\vec{R}_{R,i}(\theta(t))$ is the corresponding relative vector on the robot hand. The term $\beta_i(t)$ is an adaptive weight that dynamically adjusts the influence of this cost term based on inter-finger distance.

\textit{Adaptive Weighting ($\beta_i(t)$).}
The weight's magnitude depends on the proximity of the fingers. We first compute a normalized proximity metric $\rho_i(t) \in [0, 1]$:
\begin{equation}
    \rho_i(t) = 1 - \frac{\|\vec{R}_{H,i}(t)\| - \rho_{\min,i}}{\rho_{\max,i} - \rho_{\min,i}}
\end{equation}
where $\rho_{\min}$ and $\rho_{\max}$ are the minimum and maximum inter-finger distances determined from calibration data. This metric is then mapped through a Sigmoid function to produce the final weight:
\begin{equation}
    \beta_i(t) = \frac{1}{1 + e^{-\sigma(\rho_i(t)-\tau)}}
\end{equation}
where $\sigma$ and $\tau$ are parameters controlling the function's shape and activation threshold.

\textit{Temporal Smoothing Cost ($\mathcal{L}_{\text{smooth}}$).}
This term acts as a regularizer to penalize abrupt changes in the robot's joint angles, ensuring the motion trajectory is fluid and stable.
\begin{equation}
    \mathcal{L}_{\text{smooth}}(\theta(t)) = \| \theta(t) - \theta(t-1) \|^2
    \label{eq:smooth_loss}
\end{equation}

\subsection{Control and Data Collection}
The core of TeleDex is the integrated control and data collection pipeline, which functions as the central synchronization and serialization engine. This component is further structured into a three-layer software architecture: the Control Layer (L3), the Data Integration Layer (L2), and the Data Storage Layer (L1).

\subsubsection{Control Layer}
\label{subsec:Logic Layer}
This layer is centered around a primary ROS2 service node, which embodies the main responsibilities of the core control layer, including task manager, hardware manager, and multi-sensor timeSync manager, and data collection coordinator.

In a typical data collection workflow, an operator first specifies a hardware combination in a configuration file. Upon system launch, the corresponding ROS2 nodes in the perception component are activated, reading raw data from their respective hardware (e.g., VR devices or a master-slave arm) and publishing it as standardized ROS2 messages. 
The control layer subscribes to these standard messages. Its internal Control Strategy Scheduler then loads and applies the appropriate control algorithm based on the preset mode in the configuration. 

For instance: (1) For a high-fidelity, master-slave configuration (e.g., Agilex arm + LinkerHand L10/O6 + master arm/glove), the scheduler employs a direct joint-space mapping strategy to achieve minimal latency. (2) For a low-cost configuration with mismatched kinematics (e.g., RealMan arm + LinkerHand L10 + VR controller), the scheduler uses a hybrid control strategy: an optimization-based retargeting algorithm for the hand and Cartesian-space relative motion control for the arm. 
The standardized robot control commands generated by this layer are then published, to be subscribed to and executed by the corresponding robot driver node in the hardware component. Concurrently, a parallel data collection pipeline captures all sensor data and control commands. A time synchronization module within this pipeline aligns and validates timestamps from different sources. 

\paragraph{Data Synchronization Mechanism:}
When collecting high-quality data for imitation learning, ensuring precise temporal alignment of data from multiple heterogeneous sensors is crucial. Even a minor time deviation can lead to incorrect state-action pairing, severely impacting the learning effectiveness of downstream policies. To address this challenge, Open TeleDex implements a robust software-based timestamp synchronization (\textbf{TimeSync Manager}) and validation mechanism at the application layer, built upon ROS2's distributed clock service.  TimeSync Manager is integrated throughout the data collection process:
\begin{enumerate}
    \item \textbf{Establishment of a Global Time Base:} Upon system startup, the TimeSyncManager initializes and waits for ROS2's global clock to become stable. This ensures that all ROS2 nodes in the system share a single, synchronized time source, which is the foundation for all subsequent synchronization operations.
           
    \item \textbf{Per-Frame Multi-Source Timestamp Collection and Validation:} In each control cycle of data collection, the system performs a rigorous timestamp synchronization check:
    \begin{itemize}
        \item \textbf{Gathering}: The RealEnv interface collects timestamps for the current frame of data from all active Recorder modules. These timestamps come from various sources: camera and dexterous hand timestamps are from their ROS2 message headers, while the robotic arm's timestamp may come directly from its underlying SDK.
    
        \item \textbf{Validation}: The core function of TimeSyncManager verifies the collected set of timestamps. This validation includes two key checks:
    
        \begin{enumerate}[label=\alph*.]
            \item \textbf{Freshness Check}: Ensures all timestamps are "recent" (e.g., within 1 second of the current ROS time) to filter out stale data caused by node freezes or network issues.
            \item \textbf{Consistency Check}: Calculates the maximum difference among all timestamps in the set and checks if this difference is within a preset tolerance (e.g., 100ms). Under typical operating conditions, our system consistently keeps this difference below 70ms (tp99).
        \end{enumerate}
    \end{itemize}

    \item \textbf{Data Tagging and Storage:} The result of the validation—including a success flag, the max\_diff, and all original timestamps—is packaged as a "sync validation bundle" and stored within the current frame's observation data. This metadata is ultimately saved into the telemetry.npz file along with \textit{qpos, qvel}, etc. 
\end{enumerate}

Through TimeSync Manager, Open TeleDex achieves an active and traceable data quality assurance that goes beyond a standard rosbag. It not only strives to ensure data synchronicity at the time of collection but, more importantly, provides a synchronization quality metric for every single frame of data. During subsequent model training, researchers can easily filter out unsynchronized timestamps data with unsynchronized timestamps using the \code{sync\_validation\_is\_valid} flag, or weight the data based on the \code{max\_diff} value. This verifiable, end-to-end time synchronization loop is a key feature of Open TeleDex as a professional data collection framework for robot imitation learning.

\subsubsection{Data Integration Layer}

This layer acts as the Data Plane, specializing in acquisition, buffering, and data serialization. It is explicitly decoupled from the specific hardware and file structure.

Data acquisition node is a comprises multiple, independent Recorder modules (Arm Recorder, LinkerHand Recorder, Camera Recorder, Scalable Interface). Each Recorder subscribes to a specific hardware topic from L1, applies the globally synchronized timestamp received from the L3 TimeSync Manager, and pushes the data into the persistence buffer. 

Data persistence node is responsible for structuring and writing the buffered data to disk. It efficiently encodes and stores the synchronized, multi-modal data into a structured format (e.g., video-compressed or HDF5) with complete metadata, ready for downstream imitation learning tasks. The operator can control the start and stop of data collection via keyboard commands.

To support a new robotic arm, a developer only needs to implement a new Recorder class that adheres to the RealEnv interface, without modifying the core data collection logic. This design significantly lowers the barrier to integrating new hardware into the Open TeleDex ecosystem.

\subsubsection{Data Storage Layer}
This layer defines the final output structure and the foundational storage hierarchy, ensuring the dataset is immediately usable by learning frameworks.

Standard Storage Node enforces a consistent, hierarchical file structure. It defines the output files: Telemetry Joint Data (telemetry.npz), Cameras Data/Intrinsics (video files and calibration parameters, camera\_info.json), Episode Manifest(manifest.json), and Metadata Global Config (metadata.json).

\subsection{Hardware}

\subsubsection{Hardware Components}

\begin{figure}[!t]
    \centering
    \includegraphics[width=1.0\textwidth]{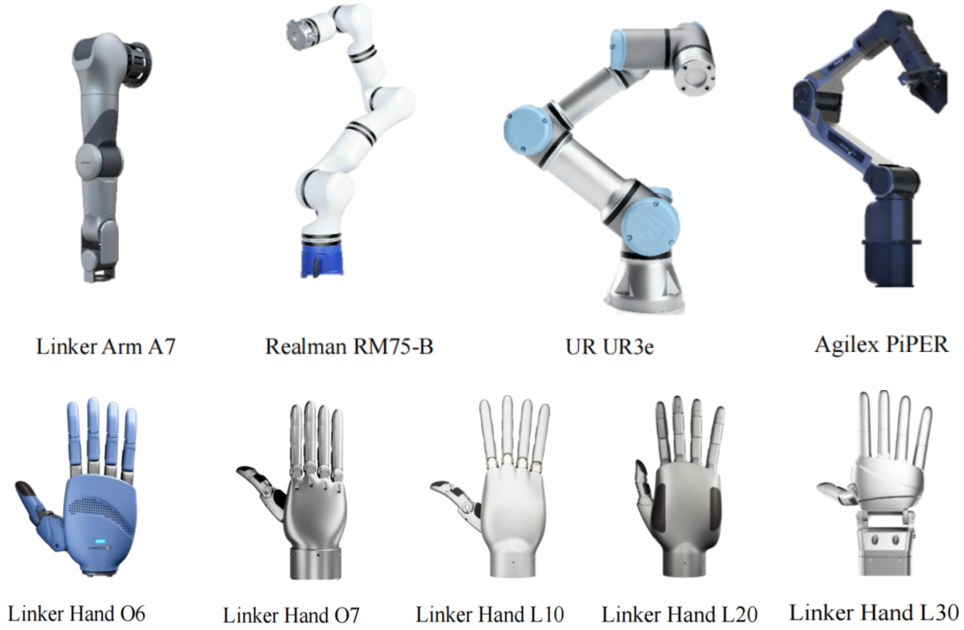} 
    \caption{Some of the Hardware Components Supported by Open TeleDex.}
    \label{fig:hardware}
\end{figure}

This component represents the physical devices and the low-level software interfaces required for execution and sensing. Fig.~\ref{fig:hardware} shows some of the hardware components supported by Open TeleDex.

\begin{enumerate}
\item Camera Driver: Standardized drivers (e.g., for RealSense) that interface directly with the camera hardware and publish synchronized RGB-D streams via ROS 2 topics. 
\item Arm Driver: Low-level controllers responsible for converting high-level velocity or joint position commands from L2 into motor signals for the mechanical arm (e.g., Piper/Linker Arm).
\item LinkerHand Driver: The specific SDK or API that provides control and state feedback for the LinkerHand, handling intricate finger articulation commands and sensor readings.
\item Sensor Data Stream: The continuous feedback loop, providing the instantaneous state (joint angles, end-effector forces, visual frames) back to the L2 Data Acquisition Node for synchronized logging and, if implemented, real-time closed-loop control.
\end{enumerate}

\subsubsection{Heterogeneous Hardware Interface}
To achieve broad hardware compatibility, we designed a core heterogeneous hardware interface. It uses software abstraction to shield the system from underlying physical hardware differences, providing a stable and unified interaction logic for upper-level applications. Its design follows two core principles: Master-side input normalization and Slave-side device abstraction.

\paragraph{Master Input Normalization: A Unified Expression of Operator Intent}

The primary challenge at the teleoperation front-end (Master side) is to handle input signals from vastly different sources and in disparate formats—for instance, data gloves output high-dimensional joint angles, VR devices provide hand keypoint poses via UDP streams, while master arms directly output their own joint states.

To mask this heterogeneity, Open TeleDex introduces an intermediary "intent parsing and normalization" step. Regardless of the input device, its raw data stream is first processed by a dedicated front-end adapter node. In our implementation, this is an independent ROS2 node named \textit{hand\_retarget}.

The core responsibility of this node is to uniformly translate these various input signals into a standardized "intent signal". This signal represents the operator's core intent and is published via a fixed ROS2 topic. Specifically:

\begin{itemize}
    \item For input from a VR controller, this node runs our hand retargeting algorithm and a Cartesian-space relative motion control strategy, driving the dexterous hand with the target hand pose and the robotic arm with the wrist keypoint data.
    
    \item For input from a data glove and a master arm, the node executes our direct joint-space mapping control strategy, mapping their poses to the standard target joint angles for the dexterous hand and robotic arm.
\end{itemize}

\paragraph{Slave Device Abstraction: A Unified Robot Environment Interface}

At the teleoperation back-end (Slave side), the challenge is to control robots of different models, degrees of freedom, and communication protocols with a single, unified logic.

To enable broad compatibility for AnyArm and AnyHand, Open TeleDex implements an environment abstraction layer named RealEnv. This design aims to provide a unified, high-level API for all physical robot hardware.

RealEnv itself is an abstract base class that defines standard methods for interacting with a robot, such as getting the current robot state and sending an action command to the robot. For each supported piece of hardware (e.g., the Agilex Piper arm or the Linker arm A7), we implement a corresponding Recorder subclass (e.g., \code{PiperRecorder}).

This Recorder subclass acts as a "device driver". It handles all low-level details of interacting with a specific hardware SDK or underlying ROS driver, while strictly adhering to the interface defined by RealEnv. For instance, when the upper-level data collection logic calls \code{get\_observation()}:

\begin{itemize}
    \item PiperRecorder queries the encoders via its SDK to get joint states.
    \item LinkerHandRecorder subscribes to the relevant ROS2 topic for the latest joint messages.
    \item RealSenseRecorder subscribes to image topics.
\end{itemize}

Despite their different internal implementations, they all ultimately return an observation dictionary with the exact same structure, containing standardized fields like qpos, qvel, and images.

Through this RealEnv abstraction, upper-level applications in the Open TeleDex can completely ignore which specific arm or hand is being used. To support a new robot in the future, a developer only needs to implement a new Recorder class following the RealEnv interface, without any modification to the core logic. This "define once, use everywhere" plug-in design is the technical cornerstone of Open TeleDex's high extensibility and hardware agnosticism.

\subsection{Summary}

Open TeleDex is engineered as a unified and general-purpose teleoperation framework designed to support a wide spectrum of hardware and application scenarios. Its key features include:

\begin{itemize}
    \item \textbf{TripleAny Compatibility}: The framework is architected to be fundamentally hardware-agnostic, supporting diverse combinations of robotic arms (\textbf{AnyArm}), dexterous hands (\textbf{AnyHand}), and master input devices (\textbf{AnyExternalDevice}), from high-fidelity master-slave systems to low-cost VR controllers.
    
    \item \textbf{Multi-Modal Data Pipeline}: Open TeleDex features a high-performance data collection pipeline, purpose-built for imitation learning. It provides robust, software-based time synchronization and validation for multi-modal data streams, including kinematic states, multi-view images, and depth data.
    
    \item \textbf{Modular, ROS2-Native Architecture}: Built entirely on ROS2, the system leverages a decoupled, three-tier architecture. This modularity, centered around a standardized RealEnv abstraction layer, vastly simplifies the integration of new hardware.
    
    \item \textbf{Flexible Control Strategies}: The framework can seamlessly switch between different control paradigms — such as direct joint-space mapping for low latency and Cartesian-space control for flexibility—based on the user's configuration.
    
    \item \textbf{AI-Ready, Structured Data Output}: All collected data is saved in a well-defined, structured format (e.g., video-compressed or HDF5) with comprehensive metadata, making it directly usable for training downstream imitation learning policies.
\end{itemize}

\begin{figure}[!t]
    \centering
    \includegraphics[width=1.0\textwidth]{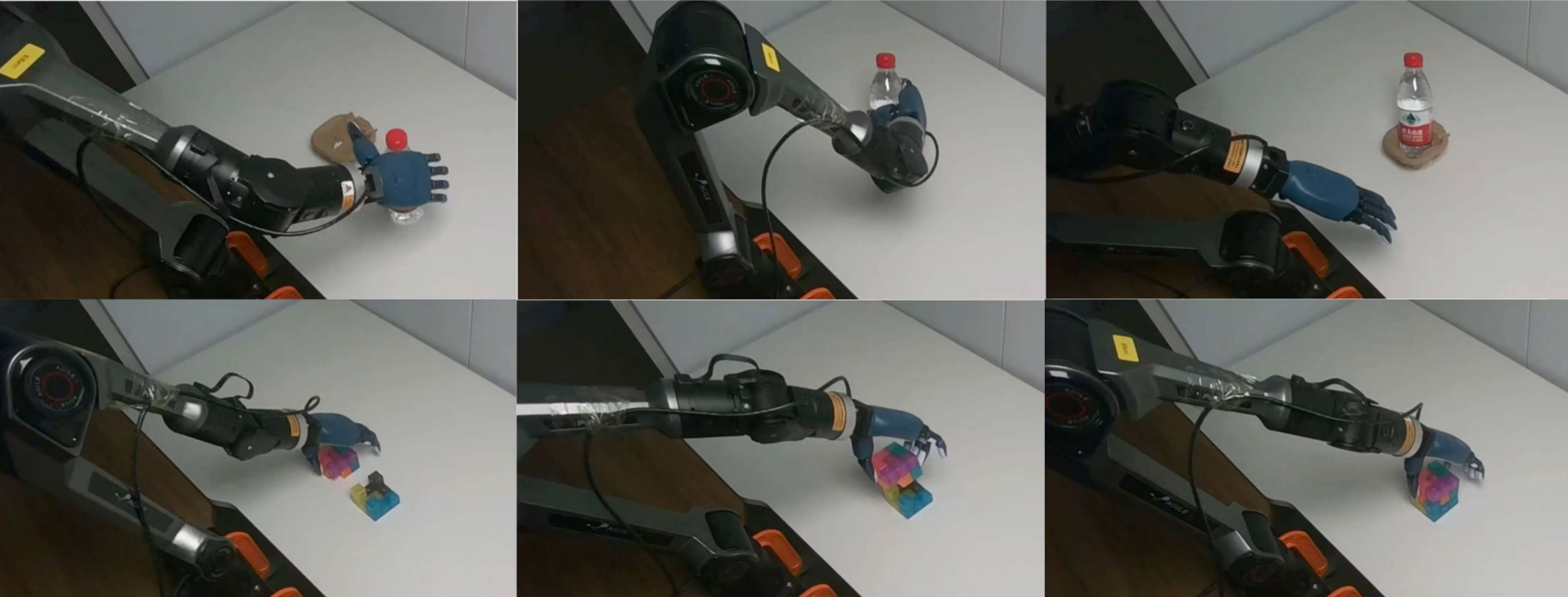} 
    \caption{This shows the task of our evaluation system. The top three images are Bottle Peg in Hole, and the bottom three images are Cube Assembly}
    \label{fig:TaskDes}
\end{figure}

\section{System Evaluation}

This section presents a rigorous evaluation of the Open TeleDex system, focusing on its capacity to support the collection of data for complex single-arm manipulation and to handle the challenges inherent in pose alignment and precise object placement. The evaluation is conducted across two benchmark tasks: Bottle Peg-in-Hole and Cube Assembly.

\subsection{Evaluation Setup and Metrics}
\subsubsection{Task Description}
\begin{description}
    \item \textbf{Task 1: Bottle Peg-in-Hole.} This task is to evaluate the system's ability to handle challenges in pose alignment when manipulating a non-rigid object. 
The operator is required to control the robot to grasp a water bottle (approximately 6cm in diameter) from the side. The robot must then insert the bottle into a square base (approximately 7cm in width and 4cm in height) placed horizontally on a tabletop. The initial horizontal distance between the two objects on the testbed is approximately 20cm.

    \item \textbf{Task 2: Cube Assembly.} To evaluate the system's capability for precise object grasping and placement, the operator is required to control the robot to pick up one half of a cube from the tabletop. The robot must then place it on top of the other half to assemble a complete cube. The initial horizontal distance between the two parts on the testbed is approximately 10cm.
\end{description}

\subsubsection{Hardware Configuration}

As show in table~\ref{tab:hardware_configurations}, the evaluation utilizes two distinct, principal robot configurations: (1) In-House Developed (TeleDex Validation Set): This configuration features our custom-built Self-Developed Robotic Arm (Linker Arm A7) paired with the Self-Developed Dexterous Hand (LinkerHand O6). (2) Commercial Off-The-Shelf (COTS) Baseline Set: This configuration utilizes mainstream, industry-standard equipment, specifically the Agilex Piper 7-DOF Arm and the LinkerHand O6 Dexterous Hand. 

\begin{table}[h!]
\centering
\caption{Hardware Configurations for System Evaluation}
\label{tab:hardware_configurations}
\begin{tabular}{lccc}
\toprule
\textbf{Experimental Group} & \textbf{Robotic Arm} & \textbf{Dexterous Hand} & \textbf{Master Device (Input)} \\
\midrule
\textbf{COTS Arm}  & Agilex Piper (6-DoF) & LinkerHand O6 & UdexReal Glove \\
\textbf{In-house Arm}  & Linker Arm A7 (7-DoF) & LinkerHand O6 & UdexReal Glove \\
\bottomrule
\end{tabular}
\end{table}

\subsubsection{Evaluation Metrics}
\label{subsubsec:metrics}

To quantitatively evaluate the performance of each system configuration, we measured a set of metrics focused on task efficiency and data quality. The key metrics are defined as follows:

\begin{description}
    \item[Average Task Duration (s)] 
    This metric measures the overall efficiency of the teleoperation system. For each successful trial, we define the task duration as the elapsed time from the moment the robot hand begins its motion towards the object to the moment the task is successfully completed (e.g., the bottle is inserted or the cube is assembled). The final value is computed by averaging the durations across all successful trials for a given task and configuration. A lower value indicates higher operational efficiency.

    \item[Synchronization Success Rate (\%)]
    This metric quantifies the reliability of our data collection pipeline. As detailed in Section~\ref{subsec:Logic Layer}, our \code{TimeSyncManager} performs a consistency check on the timestamps from all sensor streams at every single timestep. A timestep is marked as "successful" if the maximum time difference (\code{max\_diff}) among all sensor timestamps is within our preset tolerance (100ms). The Synchronization Success Rate is the percentage of successfully synchronized timesteps out of the total timesteps in an entire demonstration episode. A higher value indicates greater temporal consistency in the collected data.
    
    \item[Average Synchronization Error (ms)]
    This metric measures the temporal precision of the successfully synchronized data. It is calculated by averaging the \code{max\_diff} values (the maximum time difference between the earliest and latest sensor timestamp in a single frame) across all \textit{successfully synchronized} timesteps within an episode. This value represents the average "temporal blur" of our multi-modal data snapshots. A lower value signifies higher-quality, more precisely aligned data, which is critical for learning time-sensitive manipulation skills.
\end{description}

\paragraph{Results}
The performance results for the two baseline tasks, conducted on two distinct hardware configurations, are summarized in Table~\ref{tab:multi_group_results}. Each task was performed for 20 trials on its respective system. The table presents key metrics for task efficiency (Avg. Duration) and data quality (Sync Success and Error).

\begin{table}[h!]
\centering
\caption{Performance and Data Quality Metrics Across Different Tasks and Hardware Configurations}
\label{tab:multi_group_results}
\begin{tabular}{lcccc}
\toprule
\textbf{Configuration / Task} & \textbf{Trials} & \textbf{Avg. Duration (s)} & \textbf{Sync Success (\%)} & \textbf{Avg. Sync Error (ms)} \\
\midrule
\multicolumn{5}{l}{\textit{\textbf{COTS Arm Setup:} Agilex Piper Arm + LinkerHand O6}} \\
\quad Bottle Peg-in-Hole & 20 & 20.06 (±2.88) & 99.27 & 64.38 \\
\quad Cube Assembly       & 20 & 24.53 (±3.85) & 99.99 & 60.77 \\
\midrule
\multicolumn{5}{l}{\textit{\textbf{In-house Arm Setup:} Linker Arm A7 + LinkerHand O6}} \\
\quad Bottle Peg-in-Hole & 20 & 15.28 (±2.42) & 99.98 & 42.75 \\
\bottomrule
\end{tabular}
\end{table}

\textbf{Task Efficiency and Hardware Performance.} 
Our results first highlight the system's ability to capture task complexity. On the same hardware (COTS Arm Setup), the Cube Assembly task required a significantly longer completion time (24.53s) than the Bottle Peg-in-Hole task (20.06s), which aligns with the higher precision demands of assembly. More importantly, the results demonstrate a clear performance differentiation between hardware configurations. On the same Bottle Peg-in-Hole task, the system configured with our in-house arm was approximately \textbf{24\% faster} on average than the one with the Agilex Piper arm (COTS Arm Setup). This suggests that the tighter integration and potentially lower-level optimizations of the in-house hardware contribute to a more efficient teleoperation experience.

\textbf{Data Quality and System Robustness.}
The data quality metrics underscore the robustness of the Open TeleDex data collection pipeline. Across all tasks and hardware configurations, the Synchronization Success Rate remained above 99.2\%, indicating that our \code{TimeSyncManager} module consistently and reliably aligns multi-modal data streams within the predefined 200ms tolerance. This level of reliability is critical for generating large, clean datasets for imitation learning.

Furthermore, the \textbf{Average Synchronization Error} provides a deeper insight into the system's internal temporal precision. The in-house arm configuration not only performed tasks faster but also exhibited a significantly lower average synchronization error (42.75ms) compared to the Agilex Piper configuration (60-65ms range). A sync error of about 40-65ms, roughly equivalent to 1-2 control cycles at our 25Hz frequency, represents a solid baseline for a software-based synchronization mechanism in a complex ROS2 environment. This quantitatively demonstrates that our pipeline can maintain a consistent and measurable level of data quality, and reveals how hardware choice directly impacts the temporal fidelity of the final dataset.

\section{Conclusion and Future Work}
\label{sec:conclusion}

\subsection{Conclusion}

In this technical report, we addressed the challenge of collecting high-quality, diverse demonstration data for robot imitation learning by proposing and implementing a teleoperation framework named Open TeleDex. Our work centers on two core contributions, which have been validated through our system's design and a series of quantitative experiments.

First, we contributed a unified, device-agnostic teleoperation framework that provides an end-to-end workflow for robot learning. Through a layered, decoupled software architecture and a standardized hardware abstraction layer (RealEnv), Open TeleDex successfully breaks down the barriers between different hardware ecosystems. Crucially, this framework encapsulates the full pipeline from the synchronized collection of multi-modal data to its structured storage. The results from our data quality analysis, showing synchronization success rates consistently above 99\%, confirm that our AI-oriented data collection pipeline ensures the intrinsic quality of the collected data, providing a solid foundation for training high-performance imitation learning policies. Our experimental evaluation, using two distinct hardware configurations, demonstrated the framework's capability to seamlessly integrate and operate heterogeneous systems in practice, strongly supporting the \code{TripleAny} vision.

Second, we introduced a novel general motion mapping algorithm. This algorithm surpasses the traditional "point-to-point mapping" paradigm by generating holistic postures, which provides more natural and functional motion retargeting between morphologically different human and robot hands. Our successful demonstrations on complex tasks like Cube Assembly and Bottle Peg-in-Hole serve as a practical validation of this algorithm's effectiveness.

\subsection{Future Work}

While Open TeleDex lays a solid foundation for generalized teleoperation data collection, there is still ample room for exploration. Our future work will primarily focus on the following directions:

\paragraph{Rigorous Benchmark.}
We plan to conduct a rigorous and comprehensive performance benchmark. This involves performing standardized human-computer interaction experiments, such as Fitts' Law tests, to quantitatively measure key metrics like system throughput (IP). We will also analyze the trajectory smoothness (Jerk) of the collected data to further quantify demonstration quality. These in-depth metrics will allow for a more nuanced comparison against state-of-the-art frameworks like ALOHA and AnyTeleop.

\paragraph{Comprehensive User Experience (UX) Study.}
A comprehensive user experience (UX) study is planned to evaluate the human-in-the-loop aspects of the system. We will collect subjective metrics, such as operator workload using the NASA-TLX~\cite{hart1988development} questionnaire, to assess the intuitiveness and ergonomiscs of different hardware configurations.

\paragraph{Expansion of the Hardware Ecosystem.}
We will continue to expand the \code{TripleAny} hardware ecosystem by providing official \code{Recorder} implementations for more mainstream robotic arms, dexterous hands, and sensors, and open-sourcing them to the community to further lower the research barrier.

\paragraph{Collaborative Teleoperation.}
One of our most exciting goals is to explore and implement the "one-master-to-multiple-slaves" collaborative teleoperation mode. TeleDex's distributed architecture based on ROS2 provides a natural foundation for this objective. We plan to develop a high-level task allocation and collaborative control module to dramatically increase the diversity and efficiency of data collection for training general-purpose robot policies.

\bibliographystyle{IEEEtran}
\bibliography{reference} 


\newpage

\section*{Appendix}
\setcounter{section}{0}
\renewcommand{\thesection}{A\arabic{section}}



\section{Open TeleDex Hardware Ecosystem}
\label{app:hardware_ecosystem}

The following table lists the hardware devices that the Open TeleDex framework is designed to support, including those already integrated and those planned for future expansion. The modular architecture of Open TeleDex, particularly its \code{RealEnv} abstraction layer, facilitates the integration of new hardware by implementing a corresponding \code{Recorder} class.

\begin{table}[h!]
    \centering
    \caption{Supported and Planned Hardware for the Open TeleDex Ecosystem.}
    \label{tab:hardware_ecosystem}
    \begin{tabular}{llp{7cm}}
        \toprule
        \textbf{Category} & \textbf{Model / Type} & \textbf{Notes} \\
        \midrule
        \multicolumn{3}{l}{\textit{\textbf{Robotic Arms}}} \\
        & Agilex Piper & 6-DoF arm. \textbf{Core testbed component}. \\
        & Linker Arm A7 & 7-DoF arm. \textbf{Core testbed component}. \\
        & RealMan & Planned support. \\
        & Universal Robots (UR) & Planned support. \\
        \midrule
        \multicolumn{3}{l}{\textit{\textbf{Dexterous Hands}}} \\
        & LinkerHand O6 & 6-DoF hand. \textbf{Core testbed component}. \\
        & LinkerHand L10 & 10-DoF hand. \textbf{Core testbed component}. \\
        & LinkerHand L20 & 20-DoF hand. \textbf{Core testbed component}. \\
        & LinkerHand L30 & Planned support. \\
        \midrule
        \multicolumn{3}{l}{\textit{\textbf{Teleoperation \& Sensor Devices}}} \\
        & Master-Slave Arm / Glove & High-fidelity input device. \textbf{Integrated}. \\
        & VR Controllers (e.g., PICO, Meta Quest) & Low-cost immersive input device. \textbf{Integrated}. \\
        & UdexReal Haptic Glove & High-fidelity haptic/motion capture glove. \textbf{Integrated}. \\
        & Exoskeleton & Planned support. \\
        & Intel RealSense D455 & RGB-D Camera. \textbf{Core testbed component}. \\
        & Intel RealSense D405 & High-precision, short-range RGB-D Camera. \textbf{Verified}. \\
        \bottomrule
    \end{tabular}
\end{table}

\clearpage 

\section{Example of a Typical Task Dataset Composition}
\label{app:dataset_composition}

This appendix details the complete structure of a structured dataset generated by the Open TeleDex framework, using a single \code{episode} (\code{episode\_000015}) from a representative grasping task as an example.

Episode Metadata: this table summarizes the complete hardware and software environment, along with the collection parameters recorded for the episode, ensuring full reproducibility.

\begin{table}[h!]
    \centering
    \caption{Metadata recorded for \code{episode\_000015}.}
    \label{tab:metadata_example}
    \begin{tabular}{lll}
        \toprule
        \textbf{Category} & \textbf{Parameter} & \textbf{Value / Description} \\
        \midrule
        \textit{Episode Info} & Task Name & \code{linkerhand\_piper\_grasp} \\
        & Episode ID & \code{episode\_000015} \\
        & Session ID & \code{session\_2025-10-10\_13-46-35} \\
        & Duration (sec) & 19.64 \\
        & Timesteps & 491 \\
        \midrule
        \textit{Hardware Config} & Preset & \code{piper\_linkerhand\_o6\_single} \\
        & Arm Type & Agilex Piper (6-DoF) \\
        & Hand Type & LinkerHand O6 (6-DoF) \\
        & Total DoF & 12 \\
        \midrule
        \textit{Camera Config} & Preset & \code{intel\_d455\_single\_top} \\
        & Camera Type & Intel RealSense D455 \\
        & Position & Top-down (1280x720 @ 30fps) \\
        \midrule
        \textit{Collection Params} & Control Freq. (Hz) & 25 Hz (dt = 0.04s) \\
        & Video Codec & \code{libx264} (high quality) \\
        \midrule
        \textit{Data Structure} & Format & \code{video\_compressed} \\
        & \code{qpos} Dimension & 12 (6 arm + 6 hand) \\
        & Camera Streams & \code{cam\_top} \\
        & Tactile Sensors & \code{right\_hand} \\
        \bottomrule
    \end{tabular}
\end{table}

\end{document}

%% file: preamble.tex
\usepackage[margin=1in]{geometry} 
\usepackage{times}                
\usepackage{setspace}             
\usepackage{appendix}
\usepackage{enumitem}
\usepackage{subcaption}

\usepackage{amsmath}              
\usepackage{amssymb}              
\usepackage{siunitx}
\usepackage{authblk}

\usepackage{graphicx}             
\usepackage{booktabs}             
\usepackage{subcaption}           
\usepackage{multirow}             
\usepackage[table]{xcolor}        

\usepackage{epsfig}
\usepackage{subcaption}
\usepackage[T1]{fontenc}

\usepackage{tabularx}

\usepackage{algorithm}            
\usepackage{algpseudocode}        
\usepackage{listings}             

\definecolor{codegreen}{rgb}{0,0.6,0}
\definecolor{codegray}{rgb}{0.5,0.5,0.5}
\definecolor{codepurple}{rgb}{0.58,0,0.82}
\definecolor{backcolour}{rgb}{0.95,0.95,0.92}

\lstdefinestyle{mystyle}{
    backgroundcolor=\color{backcolour},   
    commentstyle=\color{codegreen},
    keywordstyle=\color{magenta},
    numberstyle=\tiny\color{codegray},
    stringstyle=\color{codepurple},
    basicstyle=\ttfamily\footnotesize,
    breakatwhitespace=false,         
    breaklines=true,                 
    captionpos=b,                    
    keepspaces=true,                 
    numbers=left,                    
    numbersep=5pt,                  
    showspaces=false,                
    showstringspaces=false,
    showtabs=false,                  
    tabsize=2
}
\newcommand{\code}[1]{\texttt{#1}}

\usepackage{cite}                 

\usepackage{hyperref}             
\hypersetup{
    colorlinks=true,
    linkcolor=blue,
    citecolor=green,
    filecolor=magenta,      
    urlcolor=cyan,
    pdftitle={TeleDex: A Teleoperation System for Imitation Learning based Dexterous Manipulation},
    pdfpagemode=FullScreen,
}

